# Native Language Identification using i-vector

*Ahmed Nazim Uddin[1], Md Ashequr Rahman[2], Md. Rafidul Islam[3], Mohammad Ariful Haque[4]*

Department of Electrical and Electronic Engineering
Bangladesh University of Engineering and Technology
Dhaka-1205, Bangladesh
[1] `ahmednazim.eee11@gmail.com`, [2] `ashequr186@gmail.com`, [3] `rafidul.buet@gamil.com`,
[4] `arifulhoque@eee.buet.ac.bd`,

**Abstract**

**The task of determining a speaker's native language based only on his speeches in a second language is known as Native Language Identification or NLI. Due to its increasing applications in various domains of speech signal processing, this has emerged as an important research area in recent times. In this paper we have proposed an i-vector based approach to develop an automatic NLI system using MFCC and GFCC features. For evaluation of our approach, we have tested our framework on the 2016 ComParE Native language sub-challenge dataset which has English language speakers from 11 different native language backgrounds. Our proposed method outperforms the baseline system with an improvement in accuracy by 21.95% for the MFCC feature based i-vector framework and 22.81% for the GFCC feature based i-vector framework**.

## 1. Introduction

Being the most important bio signal, speech signals transmit large volumes of information to listeners. Speech conveys not only information related to the message of the speech itself, but also about the language being spoken, and information relating to the emotion, gender, accent and identity of the speaker [1]. The goal of a Native-language Identification (NLI) system is to take all the information contained in a speaker's voice to recognize their Native Language based on their speeches in a second language. An automatic NLI system can play a role in many domains and applications, such as: speech and speaker recognition, global business and security, improvement of intelligibility of non-native speakers for deceptive speech identification etc.

A number of techniques are available in the literature that uses different approaches to develop models and classifiers for the NLI system [2][3][5][6][7]. Acoustic and prosodic features has shown notable success to discriminate among different foreign accents [4]. Phonetic knowledge also shows promising indications of the speakers' native language recognition [7][8].

Although initially introduced for speaker recognition [15], i-vector framework [9] has become very popular in the field of speech processing. Recent research found a great prospect to implement this technique in speaker verification [10], acoustic scene classification [11] [12], audio processing such as language recognition [13], music artist and genre classification [14]. In the i-vector framework speech utterances of variable length can represented by fixed-length low-dimensional feature vector. And this low dimensionality of i-vectors makes it also convenient to apply discriminative classifiers. Inspired by the success of i-vectors in different speech processing applications, we apply the same idea in the context of NLI in this work.

In this paper, we propose two different feature based i-vector framework for native language identification: MFCC feature based i-vector framework, GFCC feature based i-vector framework. Albeit the basic the i-vector framework followed by both the approaches is same, the novelty of our proposed method comes from using a distinguished feature set and PLDA scoring based classifier selection for native language evaluation. The motivation behind choosing MFCC and GFCC feature is that both of them is used very often to model the transfer function of human auditory system and such a model can be used to reflect the human perception of native language identification task. And the PLDA scoring is used for classification as it outperforms in results in similar tasks of classification for speech signal processing than the established ones like: SVM and Neural Network (NN). The performances of these frameworks are evaluated and compared on the 2016 ComParE Native language sub-challenge corpus which will be discussed in the results section.

The rest of the paper is organized as follows: in Section 2, i-vector framework for our prosed NLI system is discussed; in Section 3, the dataset and the features used to develop the frameworks are described under the experimental setup section; in Section 4, the results are presented; and in Section 5, the conclusions are derived.

## 2. i-vector framework for NLI system

In the domain of speech signal processing i-Vector subspace modeling is one of the recent methods that has become the state-of-the-art technique. This method largely provides the benefit of modeling both the intra-domain and inter-domain variabilities into the same low dimensional space retaining most of the relevant information. Figure-1 depicts the i-vector framework that is used in this dissertation for the proposed NLI system.

For extracting i-vector, speech segments are modeled using a GMM-UBM system. GMM is likely to be the most efficient model for likelihood function for speaker verification system and for forming the super-vectors (GMM and UBM) in i-vector extraction. The GMM parameters are estimated by iteratively maximizing the likelihood of the training data using an Expectation Maximization (EM) algorithm. But as the training data of a particular native language background are limited, a direct GMM modelling of a scene is inefficient. So this modelling can't be directly estimated by EM algorithm. Rather the parameters of GMM are normally adapted from a previously trained UBM by *maximum a posteriori* (MAP) adaptation whereas UBM is simply a single

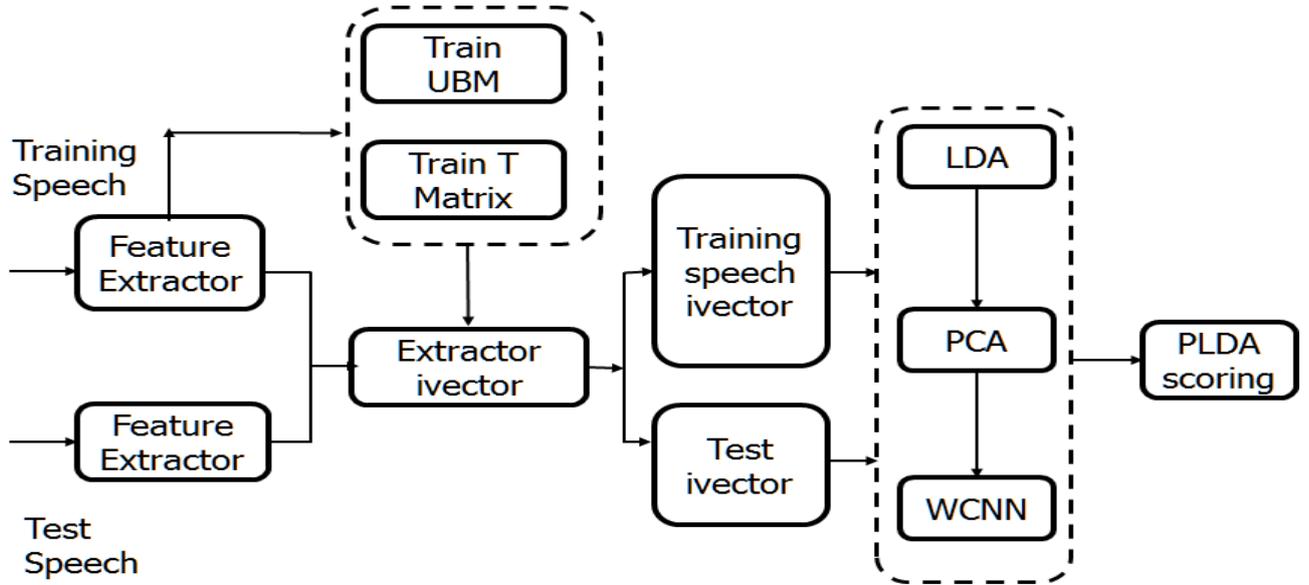

Figure1: *i-vector framework for NLI system*

GMM trained with substantial amount of data from all the native language classes at hand.

The main idea is that each speech utterance, represented by the language and channel dependent supervectors of concatenated Gaussian Mixture Model (GMM) means, can be modeled as follows:

$$M = m + Tw \qquad (1)$$

where m is the language- and channel-independent component of the mean supervector (which can be taken to be the UBM supervector), T is a matrix of bases spanning the subspace covering the important variability (both speaker- and session-specific) in the supervector space, and w is a standard normally distributed utterance dependent latent variable. For each observation sequence representing an utterance, our i-vector is the *maximum a posteriori* (MAP) point estimate of the latent variable w.

After finding i-vectors we processed them through dimensionality reduction algorithms. First, we trained the training i-vectors through Linear Discrimination Analysis [16]. The purpose is to project the higher dimensional i-vectors onto a lower-dimensional space with good class-separability in order avoid over fitting and also reduce computational costs. Then Principle Component Analysis (PCA) is done followed by LDA in which we are interested to find the directions (components) that maximize the variance in our dataset [17].

After that, we pass processed training i-vectors through Within Class Co-Variance Normalization (WCCN) [18] algorithm to reduce the within class variability in i-vector space. Then, we averaged the i-vectors for each class to get a model i-vector for each class.

After finishing the post processing and finding the model i-vectors, we used them for scoring. Here we have used PLDA scoring for classification. The class with maximum score is predicted as the class of the unknown instance.

## 3. Experimental Setup

### 3.1. Description of dataset

The performance of the proposed framework was evaluated on the ETS Corpus of Non-native spoken English which includes more than 64 hours of speech from 5,132 non-native speakers of English, with eleven different L1 backgrounds (Arabic (ARA), Chinese (CHI), French (FRE), German (GER), Hindi (HIN), Italian (ITA), Japanese (JAP), Korean (KOR), Spanish (SPA), Telugu (TEL), and Turkish (TUR)). The dataset was divided into three stratified partitions: 3,300 instances (64%) were selected as training set, 965 instances (19%) for the development set, and 867 responses (17%) used as test data. Details on this dataset is available at [19].

### 3.2. Feature extraction

From figure 1, we can see that features are extracted from the original speech signal to extract the i-vectors. Here we have developed two different feature based i-vector framework.

These features are:

(1) Mel Frequency Cepstral Coefficients (MFCCs), deltas and delta deltas.
(2) Gammatone Frequency Cepstral Coefficients (GFCCs), deltas and delta deltas

As Mel-Frequency Cepstral Coefficients (MFCC) feature is that it is the most commonly used acoustic features in speaker recognition, speech recognition, audio classifications etc. it is reasonable to take MFCC features to extract i-vectors. Gammatone Frequency Cepstral Coefficients (GFCC) is used in the same applications which outperforms MFCC features in low SNR level. So GFCC is chosen to develop the other framework. And the motivation behind the selection of the delta and delta deltas (Also known as differential and acceleration coefficients) is that is can reflect the dynamic information in the speech signals. The following parameters are used for both MFCC and GFCC feature extraction:

Table 1: Feature extraction parameter

| Parameter | Value |
| --- | --- |
| No. of bands | 26 |
| Minimum frequency | 20 Hz |
| Maximum frequency | 8000 Hz |
| No. of cepstral frequency | 20 |
| Window time | 60ms |
| Hoping time | 10ms |
| DCT type | 3 |
| Delta Width | 9 |

So for both the MFCC and GFCC, 20 cesptral coefficients are calculated augmented with 20 deltas and 20 delta deltas. Therefore, a total of 60 features is used for both i-vector based frameworks.

## 4. Results

As the labels of the test data is unknown the performance is evaluated on the development set. Here accuracy is chosen as the performance index. For the baseline system the maximum accuracy is 44.9%. A brief study on the baseline system and results with some parameter tuning is given in this section. That shows very little improvement over the baseline result. Finally results obtained from our proposed frameworks are shown where we can see that there are significant improvements from the baseline system in accuracy.

### 4.1. Study of the baseline system

The baseline is implemented using WEKA's SVM. Linear kernel with epsilon $\varepsilon = 1$ is used for the classification task. Features used for this task are extracted from the audio files using the IS13 ComParE.conf, which is included in 2.1 public release of openSMILE [20] [21]. The feature set contains 6373 static features resulting from the computation of various functionals over low-level descriptor (LLD) contours. And maximum accuracy obtained from the baseline implementation is 44.9% for complexity parameter $C = 10^{-2}$.

Then with the same feature set, the baseline is implemented in MATLAB with some parameter tuning. In this case the default value of epsilon $\varepsilon = 0.1$ is used. Changing the kernels and tuning corresponding kernel variables we get the following result:

- For linear kernel maximum accuracy: 45.3% with the value of cost parameter c = 1 which shows an improvement of about .4% than the baseline accuracy
- For RBF kernel maximum accuracy: 46.84% with the value of cost parameter c = 2.75 and gamma ɤ = -13.5 which shows an improvement of about 1.94% than the baseline accuracy.

### 4.2. Result with our proposed method

In our proposed method for both frameworks, the number of Gaussian mixtures to train the UBM, the dimension of total variability matrix (T) is tuned to get better results. And during this tuning the window length and the hop length is kept as same as the baseline system all along.

Table 2: Accuracy of MFCC feature based I-vector framework for different tuning of the model parameters

| No of gaussian Mixture | T matrix size | Accuracy (%) |
| --- | --- | --- |
| 128 | 100 | 44.18 |
|  | 200 | 58.57 |
|  | 300 | 61.17 |
| 256 | 100 | 45.55 |
|  | 200 | 61.16 |
|  | 300 | 64.77 |
| 512 | 100 | 47.54 |
|  | 200 | 62.01 |
|  | 300 | 66.85 |

Table 3: Accuracy of GFCC feature based I-vector framework for different tuning of the model parameters

| No of gaussian Mixture | T matrix size | Accuracy (%) |
|---|---|---|
| 128 | 200 | 56.14 |
| 128 | 300 | 63.34 |
| 128 | 400 | 64.77 |
| 256 | 200 | 62.65 |
| 256 | 300 | 66.25 |
| 256 | 400 | 67.53 |
| 512 | 200 | 60.59 |
| 512 | 300 | 65.77 |
| 512 | 400 | 67.71 |

From Table-2 it can be seen that the maximum accuracy of the MFCC feature based i-vector framework is 66.85% for the no of Gaussian mixture model to be 512 and the size of the T-matrix size to be 300.

From Table-2 it can be seen that the maximum accuracy of the GFCC feature based i-vector framework is 67.71% for the no of Gaussian mixture model to be 512 and the size of the T-matrix size to be 400.

So there is a relative improvement of ~1% in case of GFCC framework than that of the MFCC based one.

## 5. Conclusion

Native Language Identification of non-native speakers is a challenging problem and has diverse applications. A machine learning based technique was proposed for this task, using i-vector approach with MFCC and GFCC features. The MFCC feature based i-vector implementation outperforms the accuracy of the baseline system (accuracy: 44.9%) by 21.95% (accuracy: 66.85%). And The GFCC feature based i-vector implementation outperforms the accuracy of the baseline system by 22.81% (accuracy: 67.71%)